\documentclass{llncs}
\usepackage{llncsdoc}
\usepackage{graphicx}
\usepackage{subfigure}
\usepackage{multirow}
\usepackage{amssymb}
\usepackage{amsmath}
\usepackage{amstext}
\usepackage{amsgen}
\usepackage{textcomp}
\usepackage{amsbsy}
\usepackage{amsopn}
\usepackage{mathrsfs}
\usepackage{tikz}
\usepackage{balance}
\usepackage{hyperref}

\usepackage{todonotes}

\newcommand{\beq}{\begin{equation}}
\newcommand{\eeq}{\end{equation}}
\newcommand{\tr}[1]{{#1}^\top}
\renewcommand{\vec}[1]{\mathbf{#1}}

\newcommand{\mr}[1]{\mathrm{#1}}

\newcommand{\Real}{\mathbb{R}}

\newcommand{\GG}{\mathcal{G}}
\newcommand{\VV}{\mathcal{V}}
\newcommand{\EE}{\mathcal{E}}
\newcommand{\NN}{\mathcal{N}}
\newcommand{\LLambda}{\mathbf{\Lambda}}
\newcommand{\TTheta}{\mathbf{\Theta}}
\newcommand{\LL}{\vec{L}}
\newcommand{\II}{\vec{I}}

\newcommand{\uhat}{\widehat{\vec{u}}}
\newcommand{\xx}{\vec{x}}

\newcommand{\UU}{\vec{U}}

\newcommand{\YY}{\vec{Y}}
\newcommand{\Uhat}{\widehat{\vec{U}}}
\newcommand{\DD}{\vec{D}}
\newcommand{\AAA}{\vec{A}}
\newcommand{\mmu}{\boldsymbol{\mu}}

\begin{document}


\newcommand*{\textred}{\textcolor{red}}
\newcommand*{\textblue}{\textcolor{blue}}
\newcommand*{\textgreen}{\textcolor{green}}




 \title{Graph Convolutions on Spectral Embeddings: Learning of Cortical Surface Data}



 \author{Karthik Gopinath \and Christian Desrosiers \and Herve Lombaert }
 \institute{ETS Montreal, Canada}

\maketitle         

\begin{abstract}


Neuronal cell bodies mostly reside in the cerebral cortex.  The study of this thin and highly convoluted surface is essential for understanding how the brain works.  The analysis of surface data is, however, challenging due to the high variability of the cortical geometry.  This paper presents a novel approach for learning and exploiting surface data directly across surface domains.  Current approaches rely on geometrical simplifications, such as spherical inflations, a popular but costly process.  For instance, the widely used FreeSurfer takes about 3 hours to parcellate brain surfaces on a standard machine.  Direct learning of surface data via graph convolutions would provide a new family of fast algorithms for processing brain surfaces.  However, the current limitation of existing state-of-the-art approaches is their inability to compare surface data across different surface domains.  Surface bases are indeed incompatible between brain geometries.  This paper leverages recent advances in spectral graph matching to transfer surface data across aligned spectral domains.  This novel approach enables a direct learning of surface data across compatible surface bases. It exploits spectral filters over intrinsic representations of surface neighborhoods.  We illustrate the benefits of this approach with an application to brain parcellation.  We validate the algorithm over 101 manually labeled brain surfaces.  The results show a significant improvement in labeling accuracy over recent Euclidean approaches, while gaining a drastic speed improvement over conventional methods.

\end{abstract}

\section{Introduction}

Neuroimage analysis consists of studying functional and anatomical information over the brain geometry.  The thin outer layer of the brain cerebrum is of particular interest due to its key role in cognition, vision and perception.  Statistical frameworks on surfaces are, therefore, highly sought for studying various aspects of the brain.  For instance, variations in surface data can reveal new biomarkers as well as potential relations with disease processes \cite{Arbabshirani2017Single}.  The challenge consists of learning surface data over highly complex and convoluted surfaces.  Conventional approaches rely on geometrical simplifications, such as spherical inflation and slow mesh deformations \cite{Tustison2014Largescale}, a popular but costly process.  
For instance, the widely used FreeSurfer \cite{fischl2004:automatically} takes around 3 hours to parcellate brain surfaces by slowly deforming brain models towards labeled atlases. 
Fundamentally, current statistical frameworks exploit spatial information mostly derived from the Euclidean domain, for instance, based on vector fields, image or volumetric coordinates \cite{Zhang2011ODVBA,Hua2013Unbiased}.  Such information is highly variable across brain geometries and severally hinders the training of modern machine learning algorithms.  

State-of-the-art learning approaches \cite{Kamnitsas2017Efficient} have the potential to offer a drastic speed advantage over traditional surface-based methods, but operate on image or volumetric spaces.  Geometric deep learning \cite{Bronstein2017Geometric,Monti2017Geometric,Levie2017CayleyNets} recently proposes to use convolutional filters on irregular graphs.  This approach formulates the convolution theorem from Fourier space to spectral domains over graphs.  Furthermore, Chebyshev polynomials avoids the explicit computation of graph Laplacian eigenvectors \cite{Defferrard2016Convolutional}, and local graph filtering is made possible within small neighborhoods \cite{Monti2017Geometric}.  The main concern of these methods is their inability to compare surface data across different surface domains \cite{Bronstein2013Making,Kovnatsky2013Coupled,Ovsjanikov2012Functional,Eynard2015Multimodal}.  Laplacian eigenbases are indeed incompatible across multiple geometries.  One approach is to map local graph information onto geodesic patches and use conventional spatial convolution via template matching \cite{Masci2015Geodesic,Boscaini2016Learning}.  Geodesic patches are obtained with local parametric constructions of patches \cite{Monti2017Geometric}.  However, fundamentally, spatial representations of surface data remain defined in Euclidean spaces, for instance, using polar representations of pixels or mesh vertices.  

This paper leverages recent advances in spectral graph matching to transfer surface data across aligned spectral domains \cite{Lombaert2015Brain}.  
The transfer of spectral coordinates across domains provides a robust formulation for spectral methods that naturally handles differences across Laplacian eigenvectors, including sign flips, ordering and mixing of eigenvectors in higher frequencies.  This spectral alignment strategy was exploited to learn surface data 
\cite{Lombaert2015Spectral}, but was limited to pointwise information, ignoring local patterns within surface neighborhoods.  

Existing attempts of graph convolutions in neuroimaging \cite{Parisot2017Spectral,Ktena2017Distance} exploit single graph and rely on Euclidean representations of brain surfaces.  The learning of cortical data across multiple surfaces remain hindered by the incompatibility of spectral bases across geometries.  Our approach consists of leveraging spectral coordinates within convolutional networks. This bridges a gap between learning algorithms and geometrical representations.  To the best of our knowledge, this is the first attempt of learning surface data via spectral graph convolutions in neuroimaging.  This novel approach enables a direct learning of surface data across compatible surface bases by exploiting spectral filters over intrinsic representations of surface neighborhoods.  We illustrate the learning capabilities of this approach with an application to brain parcellation. The validation over 101 manually labeled brain surfaces \cite{Klein2017Mindboggling} shows a significant improvement of spectral graph convolutions over Euclidean approaches, from a Dice score of 59\% to 82\%.  This performance is aligned with the well established FreeSurfer algorithm \cite{fischl2004:automatically}, which scores 83\% \cite{Klein2017Mindboggling}, while gaining a drastic speed improvement, in the order of seconds.  Our contributions are multifold.  The transfer of spectral bases across domains enables the design of spectral filters in graph convolutional approaches.  Our adaptive spectral filters can consequently learn cortical surface data across multiple geometries, as well as exploit local patterns of data within surface neighborhoods.  The next section details the fundamentals of our spectral approach, followed by experiments evaluating the impact of our spectral strategy over standard Euclidean approaches for graph convolutions. 

\begin{figure}[t]
  \centering
    \includegraphics[width=\textwidth]{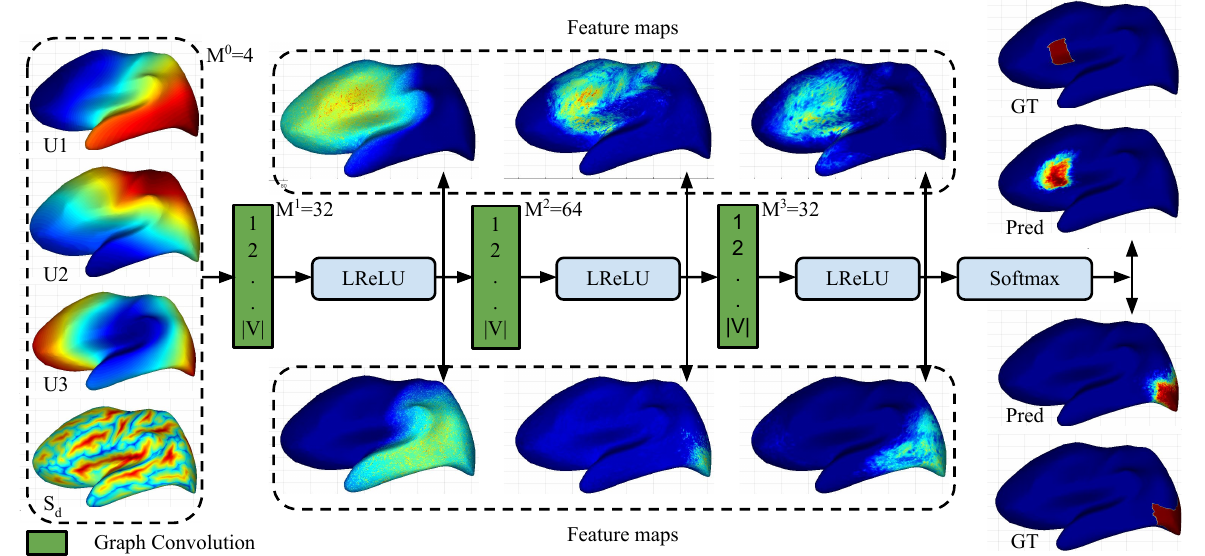}
  \caption{\textbf{Overview of the algorithm} -- On the left are inputs: Sulcal depth $S_d$, and corresponding spectral coordinates $(\uhat_1, \uhat_2, \uhat_3)$. In the middle are: Learned convolution layers, $M_l$, where sample filter responses are shown with coarse to fine geometric features. On the right are: Predicted parcel probabilities (Pred) with ground truth (GT) for two parcels. Brain surfaces are inflated for visualization.}
  \label{archi}
\end{figure}

\section{Method}


An overview of the proposed method is shown in Fig. \ref{archi}. In a first step, cortical surfaces modeled as a brain graph are embedded in a spectral manifold using the graph Laplacian operator. The spectral embedding of different brain surfaces are then aligned in the manifold using the Iterative Closest Point (ICP) algorithm. Finally, a geometric convolutional neural network (CNN) is used to map input features, corresponding to the spectral coordinates and sulcal depth of brain graph nodes, to a labeled graph.

\subsection{Spectral embedding of brain graphs}

Let $\GG = \{\VV, \EE\}$ be a brain graph defined with node set $\VV$, such that $|\VV|=N$, and edge set $\EE$. Each node $i$ has a feature vector $\xx_i \in \Real^4$ representing its 3D coordinates and sulcal depth. We map $\GG$ to a low-dimension manifold using the normalized graph Laplacian operator $\LL = \II - \DD^{-\frac{1}{2}}\AAA\DD^{-\frac{1}{2}}$, where $\AAA$ is the weighted adjacency matrix and $\DD$ the diagonal degree matrix. In this work, we define the weight between two adjacent nodes as the inverse of their Euclidean distance. Let $\LL = \UU \LLambda \tr{\UU}$ be the eigendecomposition of $\LL$, the normalized spectral coordinates of nodes are given by $\Uhat = \LLambda^{\frac{1}{2}} \UU$. 

Since the spectral embedding of $\LL$ is only defined up to an orthogonal transformation, we need to align spectral representations of different brain graphs to an arbitrary reference. Denote as $\Uhat_\mr{ref}$ the normalized spectral embedding of this reference, we align an embedding $\Uhat$ to $\Uhat_\mr{ref}$ with an iterative method based on the ICP algorithm. In this method, each node is mapped to its nearest reference node in the embedding space. The optimal orthogonal transform between matched nodes is then obtained by solving a Procrustes analysis problem. This process is repeated until convergence. 

\subsection{Graph convolution on surfaces}

We start by presenting the standard CNN model for rigid grids and then explain how this model can be extended to an arbitrary geometry. Let $\YY^{(l)} \in \Real^{N \times M_l}$ be the input feature map at convolution layer $l$ of the network, such that $y_{ip}^{(l)}$ is the $q$-th feature of the $i$-th input node. The network input thus corresponds to $\YY^{(1)}$. Assuming a 1D grid, the output feature map of layer $l$ is given by $y_{ip}^{(l+1)} = f(z_{ip}^{(l)})$ 
with
\begin{equation}\label{eq:1D-convolution-z}
    z_{ip}^{(l)} \ = \ \sum_{q=1}^{M_l} \sum_{k=-K_l}^{K_l} w_{pqk}^{(l)} \cdot y_{i+k,q}^{(l)} 
            \ + \ b_p^{(l)}.
\end{equation}
Here, $w_{pqk}^{(l)}$ are the convolution kernel weights, $b_p^{(l)}$ the bias weights of the layer, and $f$ is a non-linear activation function, for instance the sigmoid or rectified linear unit (ReLU) functions. 

To extend this fixed-grid formulation to a graph $\GG = \{\VV, \EE\}$, we denote as $\NN_i  = \{j \, | \, (i,j) \in \EE\}$ the neighbors of node $i \in \VV$. A generalized convolution operation can then be defined as
\begin{equation}\label{eq:convolution}
    z_{ip}^{(l)} \ = \ 
        \sum_{j \in \NN_i} \sum_{q=1}^{M_l} \sum_{k=1}^{K_l} 
            w_{pqk}^{(l)} \cdot y_{jq}^{(l)} \cdot \varphi(\uhat_i, \uhat_j; \, \TTheta^{(l)}_k)  \ + \ b_p^{(l)},
\end{equation}
where $\varphi(\uhat_i, \uhat_j; \TTheta_k)$ is a symmetric kernel in the embedding space with parameter $\TTheta_k$. In this work, we follow \cite{Monti2017Geometric} and use a Gaussian kernel:
\begin{equation}
  \varphi(\uhat_i, \uhat_j; \mmu_k, \sigma_k) \ = \ \exp\big(-\sigma_k \, \|(\uhat_j - \uhat_i) - \mmu_k\|^2\big).
\end{equation}

Using this formulation, we define a fully-convolutional network composed of 3 graph convolution layers with feature map sizes of $M_1 = 32$, $M_2 = 64$ and $M_3 = 32$, each one having $K_l=4$ Gaussian kernels. The size of the last layer corresponds to the number of parcels to be segmented (32 in our case). Leaky ReLU is applied after each layer to obtain filter responses: $y_{ip}^{(l)} = \max(0.01 z_{ip}^{(l)}, \, z_{ip}^{(l)})$. Since the parcels to segment are mutually exclusive, we use a softmax operation after the last graph convolution layer to obtain the parcel probabilities of each node. The softmax function is given by $\frac{exp(y_{ip}^{(l)})}{\sum_q exp(y_{iq}^{(l)})}$. Finally, cross-entropy is employed as output loss function:
\begin{equation}\label{eq:cross-entropy}
    E(\TTheta) \ = \ - \sum_{i=1}^N \sum_{c=1}^C s_{ic} \cdot \log p_{ic}(\TTheta),
\end{equation}    
where $\TTheta = \{w_{pqk}^{(l)}, \, b_p^{(l)}, \, \TTheta^{(l)}_k\}$ are the trainable network parameters,  $p_{ic}(\TTheta)$ is the output probability for node $i$ and parcel label $c$, and $s_{ic}$ is a one-hot encoding of the reference segmentation. This loss is minimized by back-propagating the error using standard gradient descent optimization.

\section{Results}

We now evaluate the performance of our contributions.  First, we highlight the advantage of moving graph learning frameworks from a conventional Euclidean domain to a Spectral domain.  Second, we assess the improvement of exploiting neighborhoods of surface data versus learning pointwise information on spectral embeddings.  Our validation is performed on Mindboggle \cite{Klein2017Mindboggling}, the largest publicly available dataset of manually labeled brain MRI.  It consists of 101 subjects collected from different sites, with cortical meshes varying from 102K to 185K vertices.  Each brain surface contains 32 manually labeled parcels.  In our experiments, the dataset was randomly split into training, validation and testing with a ratio of 70-10-20$\%$.  We further compared our results with the established FreeSurfer software.  Our implementation is in Matlab, and uses an i7 desktop computer with 16GB of RAM and an Nvidia Titan X GPU. The computation of spectral embedding and basis alignment takes roughly 15 seconds. Each training epoch takes $\sim$5-6 minutes. However, a single forward pass during testing for single subject cortical parcellation takes approximately 3 seconds.



\subsection{Euclidean versus Spectral domain}

\begin{figure}[t!]
  \centering
    \includegraphics[width=\textwidth]{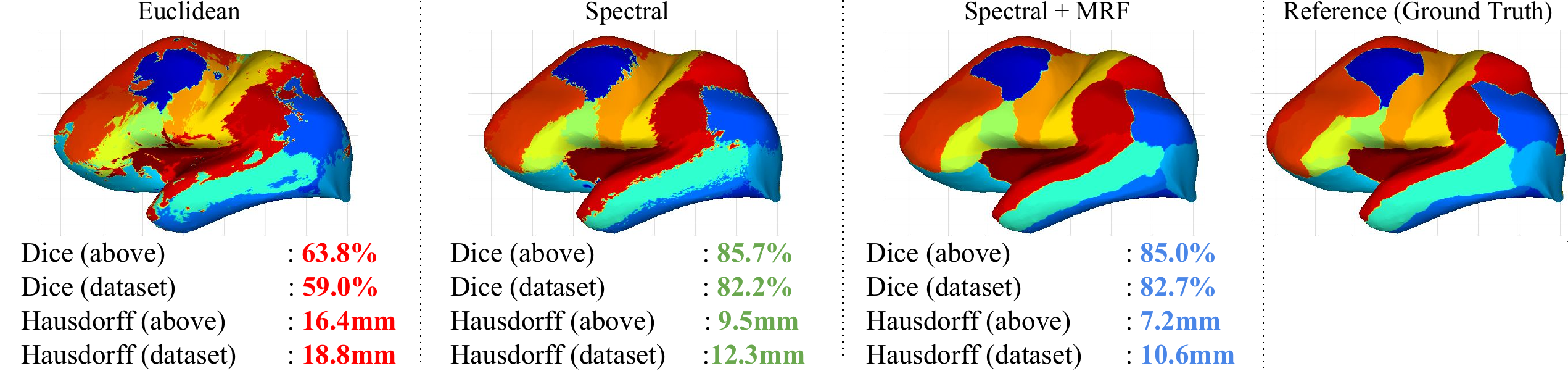}
  \caption{\textbf{Cortical Parcellation} -- (\textit{First/Left}) Learning with Euclidean coordinates, resulting in low Dice score ($59.0\%$) and inconsistent boundaries (Hausdorff distance of 18.9mm).  (\textit{Second}) Learning with Spectral coordinates, improving Dice score ($82.2\%$) and boundary regularity (12.3mm). (\textit{Third}) Spectral domain regularized with MRF, leading to consistent boundaries (10.6mm) with respect to the reference (\textit{Fourth}). The brain surface is inflated for visualization.}
 
  
  \label{eucvsspe}
\end{figure}

This experiment evaluates the improvement of moving the learning operations from the Euclidean domain to a Spectral domain.  To do so, we compare the classification accuracy on 32 cortical parcels when running our algorithm, respectively, in the Euclidean and Spectral domains. Quantitative results are measured in terms of average Dice overlap 
and Hausdorff distances \cite{Lombaert2015Spectral}. Qualitative results are shown in Fig.~\ref{eucvsspe}.  



\textbf{Euclidean domain} -- In our baseline, similarly to the latest approaches of graph convolutions networks \cite{Monti2017Geometric}, we learn from input features in the Euclidean domain. Each cortical point is represented using sulcal depth and its spatial location.
The architecture remains the same as described earlier (Fig.~\ref{archi}).  We train our network on the randomly split training set until the performance decreases on the validation set.  The average Dice overlap across all parcels in our dataset is 59.0\% ($\pm$ 20.5, min/max = 0.0/85.5\%).  The Hausdorff distance averaged across all parcels is 18.8mm. Fig.~\ref{eucvsspe} clearly illustrates the current limitation of existing graph convolutions approaches.  Spatial ambiguities are inherent in the Euclidean domain, i.e., neighboring coordinates in space may not necessarily be neighbors on surfaces.  This ambiguity may confuse training on highly convoluted surfaces such as the brain, and possibly explains the strong spatial irregularities observed in the parcels boundaries.  



\textbf{Spectral domain} -- Our contribution is to operate in a geometry-aware Spectral domain. We now learn on input features represented in the Spectral domain, where each cortical point consists of the sulcal depth with the first three spectral coordinates.
We use the same architecture and data split as before. The average Dice overlap across all parcels improves to 82.2\%($\pm$ 5.3, min/max = 69.5/95.1\%).  Details for each parcels are shown in Fig.~\ref{dice_parcels}.  The Hausdorff distance averaged across all parcels is now reduced to 12.3mm.  This is a 40\% improvement over learning in the conventional Euclidean domain.  The qualitative results of Fig.~\ref{eucvsspe} show that our cortical parcellation is almost similar to the manual parcellation.  The boundary, however, is irregular and requires further regularization.  

As an illustration of further refinement, we use Markov random field (MRF) regularization for both Euclidean and Spectral outputs. Toward this goal, we apply a standard graphcut algorithm \cite{graphcut} with minus-log parcel probabilities as unary potentials and the Potts model for defining binary potentials. MRF regularization further improves the overall classification accuracy from 68.2\% to 71.1\% in the Euclidean domain, and from 84.4\% to 85.4\% in the Spectral domain. Similar improvement is observed in terms of Hausdorff distance, with a reduction from 18.8mm to 15.9mm in the Euclidean domain, and from 12.3mm to 10.6mm in the Spectral domain.  The Dice overlap scores remain, however, similar after regularization, perhaps due to a correct initial overlap of parcels by our algorithm.  


  

  

\subsection{Pointwise versus neighborhood information}

\begin{figure}[t!]
  \centering
    \includegraphics[width=\textwidth]{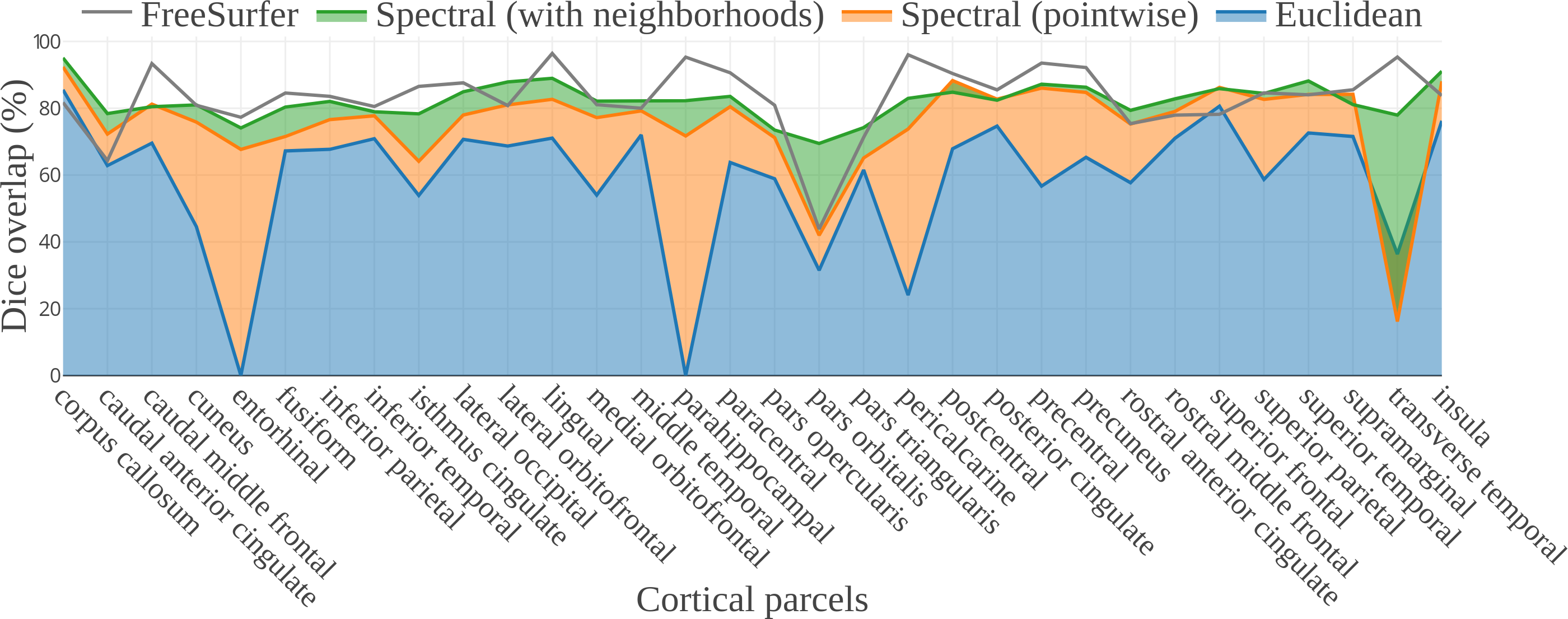}
  \caption{\textbf{Performance Evaluation} -- Dice scores for all 32 cortical parcels across the dataset when learning with: (\textit{Blue}) Euclidean, (\textit{Orange}) pointwise spectral, and (\textit{Green/Ours}) neighborhoods of spectral coordinates.  (\textit{Black}) FreeSurfer is shown for comparison.  Standard deviations of Dice scores for all parcels: (Euclidean) 20.5\%, (pointwise spectral) 14.3\%, (FreeSurfer) 10.3\%, (spectral neighborhoods) 5.3\%, which is the least variable.}
  \label{dice_parcels}
\end{figure}


In this experiment, we illustrate the benefits of exploiting neighborhoods of surface data over pointwise information in a spectral domain.  To follow standard approaches of learning pointwise surface data \cite{Lombaert2015Spectral}, we remove neighboring connections by setting our weighted adjacency matrix as an identity matrix. In addition, we disable the update of kernel parameters in convolutional layers. Training on pointwise information 
results in an average Dice overlap of 75.6\%($\pm$ 14.3\%, min/max = 16.2/92.3\%).  As a reminder, training on neighborhoods results in 82.2\% of Dice score.  Exploiting neighborhoods of surface data thus improves the performance of our algorithm.  A closer look to the performance scores for each parcel (Fig.~\ref{dice_parcels}) also reveals a general improvement when exploiting neighborhoods over pointwise surface data.  Results for all parcels are also at par with FreeSurfer's performance.  


In the MindBoggle dataset \cite{Klein2017Mindboggling}, the manual parcellations were created by experts who corrected parcels initially generated from FreeSurfer. The agreement between manually corrected parcels and FreeSurfer is 83.2\% ($\pm10.3\%$) of Dice score.  Our algorithm produces a 82.2\% ($\pm5.3\%$) Dice score with respect to manual parcellation, while gaining a significant improvement in computation time, from 3 hours to 18 seconds for processing one subject. Our graph convolutional framework also results in a stable performance when compared to an Euclidean approach or to FreeSurfer, who respectively have a standard deviation on Dice scores of 20.5\% and 10.3\% across all parcels.  Our spectral approach varies by 5.3\%.  

\section{Conclusion} 
\label{sec:diss_con}

This paper presented a novel framework for learning surface data via spectral graph convolutions.  The algorithm leverages recent advances in spectral matching to enable the comparison of surface data across different surface domains.  Our experiments illustrated the benefits of our approach with an application to cortical surface parcellation.  This is a particularly challenging problem where current graph convolution approaches remain limited by the inability to compare surface data across brain geometries.  This typically results in spatial irregularities of parcel boundaries as illustrated in Fig.~\ref{eucvsspe}.  
%
%
By capturing the geometry of the spectral manifold, the proposed method can improve the parcellation accuracy to a Dice score of $82.2\%$, from $59.0\%$ with graph convolutions in the Euclidean space. The performance of our method is comparable to state-of-the-art approaches for cortical parcellation, however, it reduces the computation time by an order of magnitude (18 seconds vs hours for FreeSurfer). 
While the potential of our method was demonstrated on cortical parcellation, it can be applied to other analyses of surface data. For instance, our framework has direct impact in regression problems that involve predictions of cortical thickness over time, potentially leading to new families of geometry-based biomarkers for neurological disorders. 

%
%

\bibliographystyle{splncs}
\bibliography{Reference}

\begin{thebibliography}{10}

\bibitem{Arbabshirani2017Single}
Arbabshirani, M.R., Plis, S., Sui, J., Calhoun, V.D.:
\newblock Single subject prediction of brain disorders in neuroimaging:
  Promises and pitfalls.
\newblock NeuroImage (2017)

\bibitem{Tustison2014Largescale}
Tustison, N.J.,  et~al.:
\newblock Large-scale evaluation of {ANTs} and {FreeSurfer} cortical thickness
  measurements.
\newblock NeuroImage (2014)

\bibitem{fischl2004:automatically}
Fischl, B.,  et~al.:
\newblock Automatically parcellating the human cortex.
\newblock C. Cortex (2004)

\bibitem{Zhang2011ODVBA}
Zhang, T., Davatzikos, C.:
\newblock {ODVBA}: optimally-discriminative voxel-based analysis.
\newblock IEEE TMI (2011)

\bibitem{Hua2013Unbiased}
Hua, X.,  et~al.:
\newblock Unbiased tensor-based morphometry: improved robustness and sample
  size estimates for {AD} clinical trials.
\newblock NeuroImage (2013)

\bibitem{Kamnitsas2017Efficient}
Kamnitsas, K.,  et~al.:
\newblock Efficient multi-scale {3D} {CNN} with fully connected {CRF} for
  accurate brain lesion segmentation.
\newblock Medical image analysis (2017)

\bibitem{Bronstein2017Geometric}
{Bronstein}, M., {Bruna}, J., {LeCun}, Y., {Szlam}, A., {Vandergheynst}, P.:
\newblock Geometric deep learning: Going beyond euclidean data.
\newblock IEEE Signal Processing (2017)

\bibitem{Monti2017Geometric}
Monti, F., Boscaini, D., Masci, J., Rodol\`{a}, E., Svoboda, J., Bronstein,
  M.M.:
\newblock Geometric deep learning on graphs using mixture model {CNNs}.
\newblock In: CVPR. (2017)

\bibitem{Levie2017CayleyNets}
Levie, R., Monti, F., Bresson, X., Bronstein, M.:
\newblock {CayleyNets}: Graph convolutional neural networks with complex
  rational spectral filters.
\newblock arxiv (2017)

\bibitem{Defferrard2016Convolutional}
Defferrard, M., Bresson, X., Vandergheynst, P.:
\newblock Convolutional neural networks on graphs with fast localized spectral
  filtering.
\newblock In: NIPS. (2016)

\bibitem{Bronstein2013Making}
Bronstein, M., Glashoff, K., Loring, T.:
\newblock Making laplacians commute.
\newblock CoRR (2013)

\bibitem{Kovnatsky2013Coupled}
Kovnatsky, A., Bronstein, M.M., Bronstein, A.M., Glashoff, K., Kimmel, R.:
\newblock Coupled quasi-harmonic bases.
\newblock In: Computer Graphics Forum. (2013)

\bibitem{Ovsjanikov2012Functional}
Ovsjanikov, M., Ben-Chen, M., Solomon, J., Butscher, A., Guibas, L.:
\newblock Functional maps: A flexible representation of maps between shapes.
\newblock In: SIGGRAPH. (2012)

\bibitem{Eynard2015Multimodal}
Eynard, D.,  et~al.:
\newblock Multimodal manifold analysis by simultaneous diagonalization of
  {Laplacians}.
\newblock IEEE PAMI (2015)

\bibitem{Masci2015Geodesic}
Masci, J., Boscaini, D., Bronstein, M., Vandergheynst, P.:
\newblock Geodesic convolutional neural networks on {Riemannian} manifolds.
\newblock In: ICCV-3dRR. (2015)

\bibitem{Boscaini2016Learning}
Boscaini, D., Masci, J., Rodol\`{a}, E., Bronstein, M.:
\newblock Learning shape correspondence with anisotropic convolutional neural
  networks.
\newblock In: NIPS. (2016)

\bibitem{Lombaert2015Brain}
Lombaert, H., Arcaro, M., Ayache, N.:
\newblock Brain transfer: Spectral analysis of cortical surfaces and functional
  maps.
\newblock In: IPMI. (2015)

\bibitem{Lombaert2015Spectral}
Lombaert, H., Criminisi, A., Ayache, N.:
\newblock Spectral forests: Learning of surface data, application to cortical
  parcellation.
\newblock In: MICCAI. (2015)

\bibitem{Parisot2017Spectral}
Parisot, S.,  et~al.:
\newblock Spectral graph convolutions for population-based disease prediction.
\newblock In: MICCAI. (2017)

\bibitem{Ktena2017Distance}
Ktena, S.I.,  et~al.:
\newblock Distance metric learning using graph convolutional networks:
  Application to functional brain networks.
\newblock In: MICCAI. (2017)

\bibitem{Klein2017Mindboggling}
Klein, A.,  et~al.:
\newblock Mindboggling morphometry of human brains.
\newblock PLOS Bio (2017)

\bibitem{graphcut}
Boykov, Y., Kolmogorov, V.:
\newblock An experimental comparison of min-cut/max- flow algorithms for energy
  minimization in vision.
\newblock IEEE PAMI (2004)

\end{thebibliography}

\end{document}